\documentclass[sigconf]{acmart}

\AtBeginDocument{%
  }

\copyrightyear{2023}
\acmYear{2023}
\setcopyright{acmlicensed}\acmConference[AHs '23]{Augmented Humans Conference}{March 12--14, 2023}{Glasgow, United Kingdom}
\acmBooktitle{Augmented Humans Conference (AHs '23), March 12--14, 2023, Glasgow, United Kingdom}
\acmPrice{15.00}
\acmDOI{10.1145/3582700.3582722}
\acmISBN{978-1-4503-9984-5/23/03}

\usepackage{caption}
\usepackage[subrefformat=parens]{subcaption}
\begin{document}

\title{AIx Speed: Playback Speed Optimization Using Listening Comprehension of Speech Recognition Models}

\author{Kazuki Kawamura}
\affiliation{
\institution{The University of Tokyo, Tokyo, Japan}
\country{}
}
\affiliation{
\institution{Sony CSL Kyoto, Kyoto, Japan}
\country{}
}
\email{kwmr@acm.org}

\author{Jun Rekimoto}
\affiliation{
\institution{The University of Tokyo, Tokyo, Japan}
\country{}
}
\affiliation{
\institution{Sony CSL Kyoto, Kyoto, Japan}
\country{}
}
\email{rekimoto@acm.org}

\begin{abstract}
Since humans can listen to audio and watch videos at faster speeds than actually observed, we often listen to or watch these pieces of content at higher playback speeds to increase the time efficiency of content comprehension. 
To further utilize this capability, systems that automatically adjust the playback speed according to the user's condition and the type of content to assist in more efficient comprehension of time-series content have been developed. 
However, there is still room for these systems to further extend human speed-listening ability by generating speech with playback speed optimized for even finer time units and providing it to humans. 
In this study, we determine whether humans can hear the optimized speech and propose a system that automatically adjusts playback speed at units as small as phonemes while ensuring speech intelligibility. 
The system uses the speech recognizer score as a proxy for how well a human can hear a certain unit of speech and maximizes the speech playback speed to the extent that a human can hear.
This method can be used to produce fast but intelligible speech. 
In the evaluation experiment, we compared the speech played back at a constant fast speed and the flexibly speed-up speech generated by the proposed method in a blind test and confirmed that the proposed method produced speech that was easier to listen to.
\end{abstract}

\keywords{video information, playback speed, deep neural network, speech recognition, self-supervised learning}

\begin{teaserfigure}
  \includegraphics[width=1.0\textwidth]{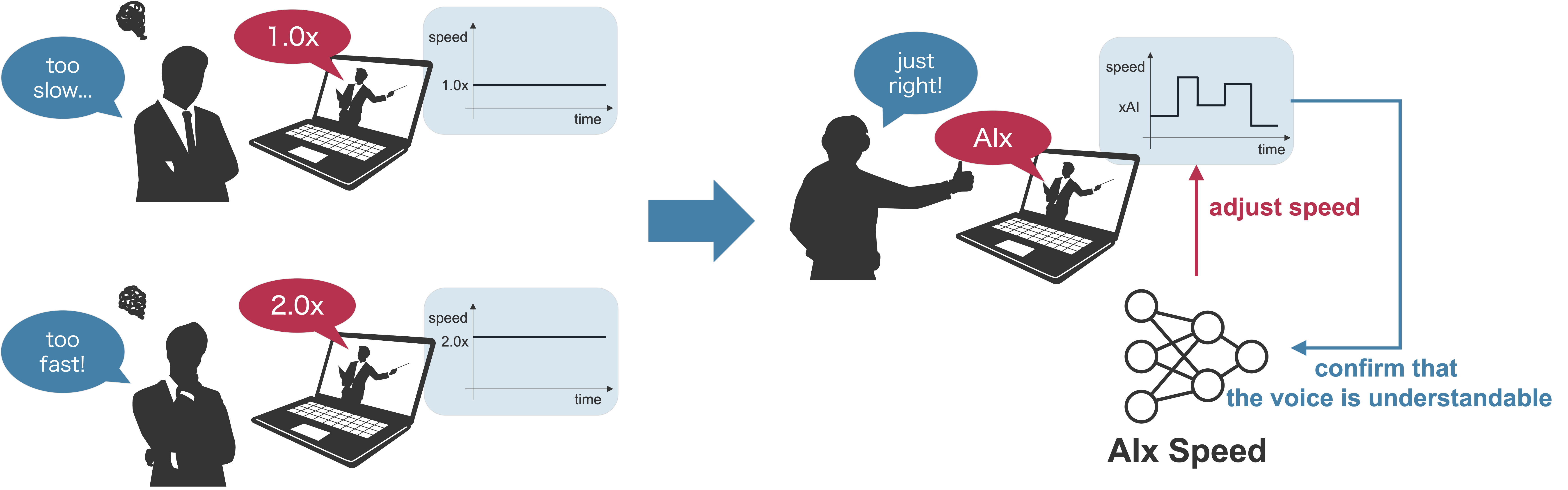}
  \caption{AIx Speed optimizes the playback speed of a video in units as small as phonemes. The system maximizes the speed of audio to the extent that humans can hear it. Users can watch videos at a comfortable speed without having to adjust the playback speed.}
  \label{fig:overview}
\end{teaserfigure}

\maketitle

\section{INTRODUCTION}
With the widespread use of video distribution services, people are increasingly watching videos for various purposes, including information gathering, learning, and entertainment. 
Humans are capable of understanding naturally observed phenomena at rates faster than the original. 
Therefore, when watching video or listening to audio, we often increase the playback speed of the content to understand more content in a shorter amount of time. 
Existing research has shown that when watching videos for learning, under certain conditions, differences in video playback speed do not affect learning effectiveness and may even improve performance~\cite{Nagahama2017, Lang2020, Dillon2022}. 
It has also been reported that 30\% to 80\% of users prefer to watch dramas at high speed, although this varies from country to country~\cite{Duan2019}.

There are many advantages to listening to video and audio at high speeds.
As a result, many methods have been proposed to automatically adjust the playback speed according to the structure of the content and the condition of the user, to further enhance the human ability to listen at high speeds.
They are widely studied as video summarization~\cite{apostolidis2021video} and audio summarization~\cite{vartakavi2021audio}.  
There are two basic strategies to these methods. 
The first is to scan the user's intentions and behavior and leave only the parts that need to be viewed or adjust the playback speed in proportion to the user's concentration level~\cite{Kurihara2011, Kawamura2014}. 
The second is to speed up unnecessary parts of the content (i.e., parts that do not contain speech) or slow down parts that contain speech~\cite{Kayukawa2018, Higuchi2017, Song2015, Zhang2020}. 
However, these methods do not explicitly model whether the resulting speech at different playback speeds is intelligible to humans, so it is unclear whether a wide range of speech types can be made intelligible to users. 
In addition, these systems vary the playback speed for each large chunk of speech, which leaves room for adjustment regarding playback speed for smaller units of time.

Therefore, we propose AIx Speed, a system that adjusts audiovisual output speed while maintaining intelligibility by measuring speech intelligibility after playback speed is increased.
As shown in Fig.~\ref{fig:overview}, this system flexibly optimizes the playback speed in a video at the phoneme level. 
By utilizing the listening ability of a neural network--based speech recognition model, which is said to rival human performance~\cite{Xiong2016}, the system simultaneously maximizes the video playback speed and the speech recognition rate after changing the playback speed. 
In this paper, the validity of using speech recognizers as a proxy for evaluating human listening performance was tested through the correlation of changes in human and speech listening performance when speed is increased.
We also whether speech where the playback speed is controlled at the phoneme level, as generated by the proposed method, or speech that is played at a constant speed and a high rate is easier for humans to listen to. 
The results showed that the utterances generated by the proposed method were easier for humans to understand. 
Furthermore, the experimental results confirmed that the speech of non-native speakers can be transformed into speech that is easier to understand for native speakers by speeding up the speech with AIx Speed. 
In summary, the proposed method not only supports the improvement of human speed--listening ability, but also improves the intelligibility of speech by generating speech with adjustable playback speeds that consider the balance between playback speed and speech intelligibility.

The contributions of this paper are summarized as follows.
\begin{quote}
\begin{itemize}
  \item Demonstrates that speech recognizers can be a substitute for human listening performance assessment.
  \item Proposes a method to increase playback speed while maintaining speech intelligibility at the phoneme level.
  \item Improves the intelligibility of speech for non-native speakers by optimizing speech rate at the phoneme level.
\end{itemize}
\end{quote}

\begin{figure}[t]
  \includegraphics[width=0.5\textwidth]{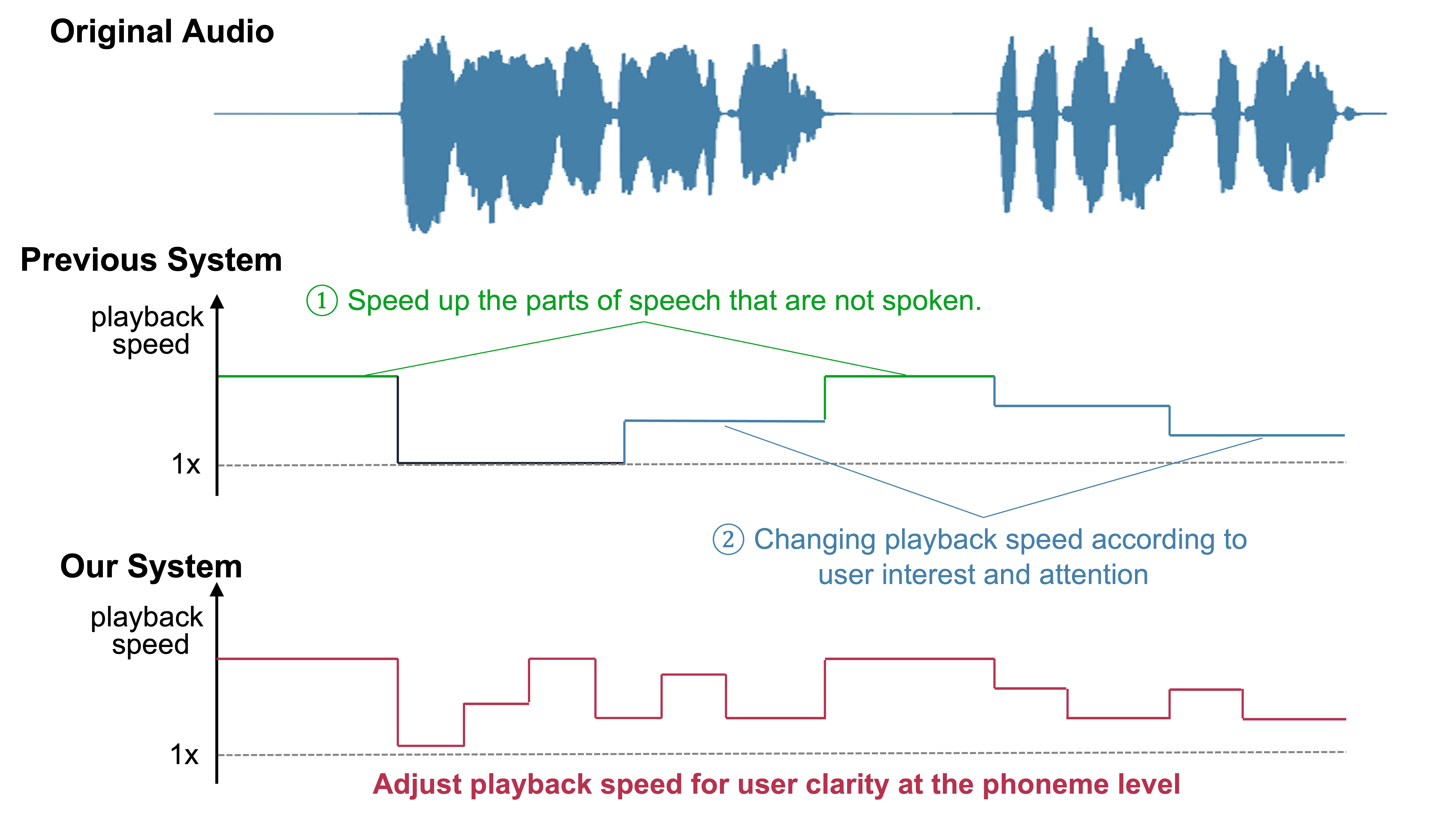}
  \caption{Differences between existing methods and ours. Existing methods speed up the playback speed of silent or unimportant parts of the audio source and slow down the playback speed of important parts of the audio source or when the user is not concentrating on them. Our method, on the other hand, adjusts the playback speed of the sound source at finer intervals according to the audibility of the speaker's speech in the sound source.}
  \label{fig:diff_related_work}
\end{figure}

\section{RELATED WORK}
There are two main methods for adjusting video playback speed to improve the time efficiency of video viewing.
The first is to remove unnecessary portions by focusing on the content, and the second is to retain the necessary portions based on user interaction.
The former removes portions that do not have audio, which is accomplished using systems such as CinemaGazer~\cite{Kurihara2011}, or portions of sports games that are not highlights of the game~\cite{Kawamura2014}. 
The latter has been studied extensively, especially in the field of human--computer interaction (HCI), and adjusts the playback speed based on the user's behavior. 
For example, SmartPlayer~\cite{Cheng2009} learns the optimal playback speed based on a user's past viewing history. 
There are also technologies that allow a user to make a rough selection in advance of what AIx Speed considers important and then fast-forward the rest of the video~\cite{Kayukawa2018, Higuchi2017}. 
Others monitor the user's movements and adjust the playback speed according to the user's level of concentration~\cite{Song2015} and comprehension~\cite{Zhang2020, Nishida2022}. 
These technologies have the advantage of tracking the optimal playback speed for each user, but they cannot reflect important factors such as the intelligibility of the conversation in the video and its changes in the playback speed.

These methods do not explicitly model whether the resulting speech is intelligible to humans when the playback speed is varied.
Therefore, speeding up the playback of various types of audio and video while making the audio understandable to the user is still an open problem.
In addition, these systems vary the playback speed for large chunks of speech, which leaves room for adjustment regarding the playback speed for smaller units of time.
In this respect, the proposed system can adjust the playback speed in finer phoneme units and can also generate speech that is easier for the user to understand.
The differences between the proposed system and the existing systems are shown in Fig.~\ref{fig:diff_related_work}.

\begin{figure*}[h]
  \begin{minipage}[b]{1.0\columnwidth}
    \centering
    \includegraphics[width=\columnwidth]{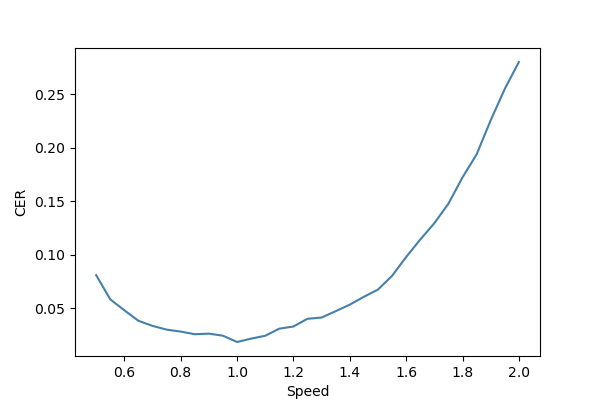}
    \subcaption{CER of ML models}
  \end{minipage}
  \hspace{0.04\columnwidth} 
  \begin{minipage}[b]{1.0\columnwidth}
    \centering
    \includegraphics[width=\columnwidth]{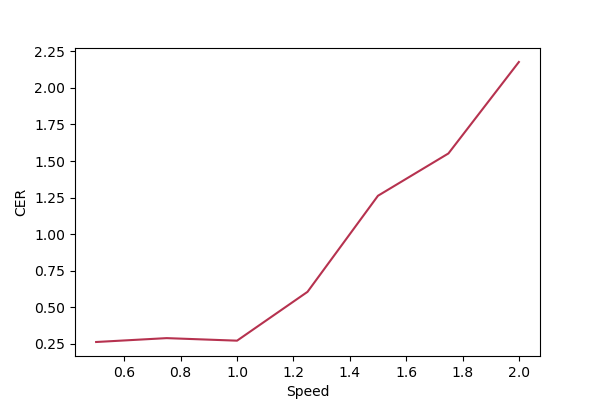}
    \subcaption{CER of humans}
  \end{minipage}\\
    \begin{minipage}[b]{1.0\columnwidth}
    \centering
    \includegraphics[width=\columnwidth]{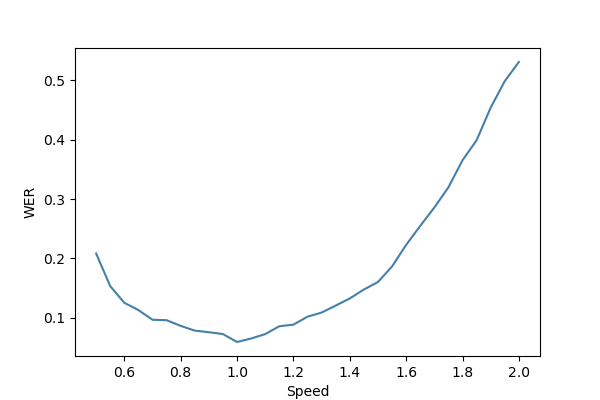}
    \subcaption{WER of ML models}
  \end{minipage}
  \hspace{0.04\columnwidth} 
  \begin{minipage}[b]{1.0\columnwidth}
    \centering
    \includegraphics[width=\columnwidth]{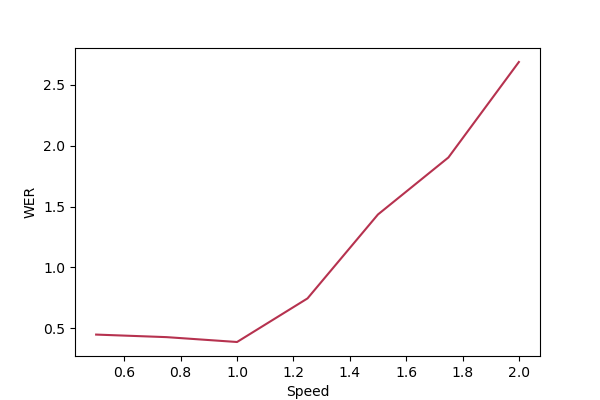}
    \subcaption{WER of humans}
  \end{minipage}
  \caption{Comparison of playback speed and listening comprehension in machine learning (ML)-based speech recognizers and humans. Transitions in human and speech recognition model transcription performance at speeds greater than 1x are correlated by a coefficient of 0.99.}
  \label{fig:pilot_study}
\end{figure*}

\section{Pilot Survey}
The purpose of this study is to automate the maximization of speed to the extent that speech is understandable. 
To this end, we hypothesize that as the speed increases, the recognition performance of both humans and speech recognizers will decrease in the same way.
If this hypothesis is correct, we can evaluate how well a human can hear when playback speed is increased using a speech recognizer instead of a human.
Several attempts have been made to evaluate human hearing with speech recognizers in this way. For example, it has been shown that the results of human mean opinion score (MOS) listening tests correlate with the results of the speech recognition-based MOS estimation method introduced in \cite{Jiang2002}, and in \cite{Lionel2017}, the understanding and comprehension scores of a listener with simulated age-related hearing loss were highly correlated speech recognition-based system.
A similar hypothesis has been used in speech learning support research to evaluate a learner's speech ability based on speech recognition performance~\cite{Cristian2021}. 
In other words, if a speech recognizer can recognize speech, it judges that the person speaks well. 
However, the relationship between the machine learning model and the ability to understand human speech when the playback speed is varied, which is the focus of this study, has not been evaluated. Therefore, we first investigated whether this hypothesis is true.

This study compared human listening performance and speech recognition performance for speech at 0.25, 0.5, 0.75, 1.0, 1.25, 1.5, 1.75, and 2.0x playback speeds.
To measure human listening performance, speech data of English sentences were prepared at each playback speed, and the subjects were asked to transcribe the data. 
The target English sentences were selected from LibriSpeech~\cite{Vassil2015}, a large English speech corpus created for the development and evaluation of automatic speech recognition systems.
This corpus contains over 1,000 hours of audiobook readings and transcriptions.
The participants were 140 English speakers who lived in the United States and had graduated from a US high school.
All participants were recruited from the Amazon Mechanical Turk and were compensated for their time.
Each subject was given 15 English sentences that had been randomly sped up by 0.5, 0.75, 1.0, 1.25, 1.5, 1.75, or 2.0x and asked to transcribe them.
The speech recognizer transcription data was collected by inputting 15 English sentences at each playback speed into a Wav2Vec2-based speech recognition model, similar to the human performance evaluation~\cite{Alexei2020}.

Figure~\ref{fig:pilot_study} shows a graph of the change in listening performance of the human and machine learning models when the playback speed was changed.
In these figures, the horizontal axis is the playback speed, and the vertical axis is the recognition performance. 
Recognition performance was evaluated using character error rate (CER) and word error rate (WER), which are commonly used in speech recognition (the lower these values are, the better). 
The WER was calculated as follows:
{\footnotesize
\begin{align*}
\cfrac{\rm (\#~of~inserted~words + \#~of~replaced~words + \#~of~deleted~words)}{\rm (\#~of~correct~words)},
\end{align*}
}
and CER was calculated as
{\footnotesize
\begin{align*}
\cfrac{\rm (\#~of~inserted~characters + \#~of~replaced~characters + \#~of~deleted~characters)}{\rm (\#~of~correct~characters)}.
\end{align*}
}
For both the human and machine learning models, listening performance decreased as playback speed increased from 1.0x. 
For playback speeds greater than 1.0x, the correlation coefficient between the change in listening performance for the human and machine learning models was 0.9977.
On the other hand, when the playback speed was slowed down from 1.0x, the machine learning model showed a decrease in recognition performance, but the decline barely observable for humans. 
In particular, while the performance of WER decreased slowly, the performance of CER showed almost no decrease. 
This indicates that human recognition performance on a character-by-character basis does not drop nearly as much when listening to slowed speech. 
Thus, when the playback speed increased, the recognition performance of the human and machine learning models decreased similarly, but they exhibited different behavior when the playback speed was decreased. 
However, since this study focuses on increasing the playback speed of speech, the difference in behavior when the playback speed is slower than 1.0x has no effect. 
Therefore, by taking advantage of the fact that listening performance decreases as playback speed is increased for humans and speech recognition models alike, we replaced human listening performance with speech recognition performance to develop the desired system.

\section{AIx Speed}  
\begin{figure}[t]
  \includegraphics[width=0.5\textwidth]{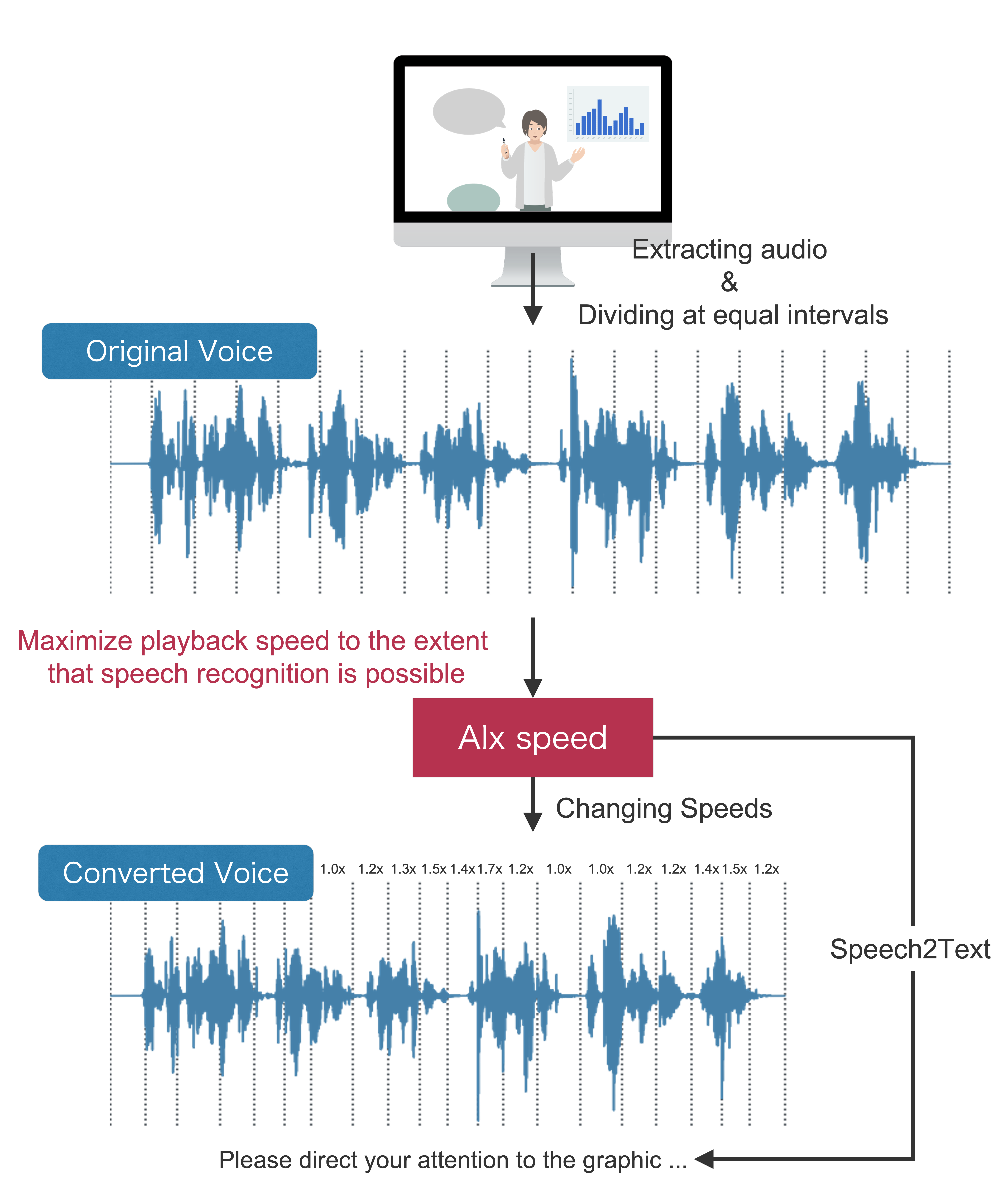}
  \caption{The work process of AIx Speed: \textcircled{\scriptsize 1}~Extract human voices from the target video. \textcircled{\scriptsize 2}~Divide the voices into specified equal intervals. \textcircled{\scriptsize 3}~Calculate the optimal playback speed for each divided voice. \textcircled{\scriptsize 4}~Change each voice to the fixed playback speed and synthesize it into one voice. \textcircled{\scriptsize 5}~Perform speech recognition on the synthesized voice to confirm that the resulting voice is understandable.}
  \Description{}
  \label{fig:aix_speed}
\end{figure}

\begin{figure*}[t]
  \includegraphics[width=0.95\textwidth]{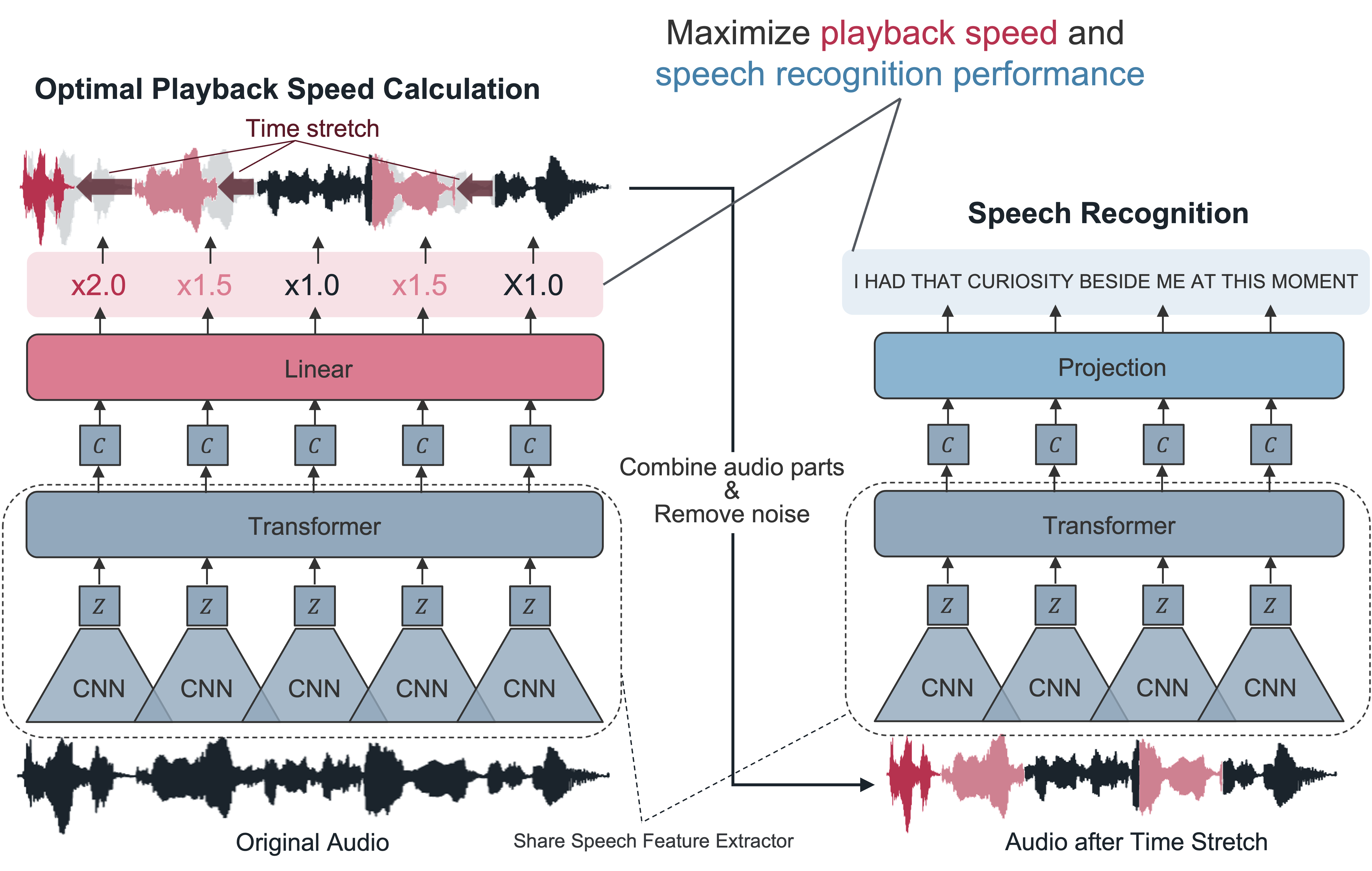}
  \caption{AIx Speed architecture: Our system simultaneously optimizes the playback speed regulator (left) and the speech recognizer (right). By doing so, we can maximize the playback speed to the extent that the model can recognize.}
  \Description{}
  \label{fig:architecture}
\end{figure*}

AIx Speed increases the speed as much as possible, as long as the user can understand it.
This system allows users to watch videos in a time-efficient manner without having to adjust the playback speed for each video. 
In addition, the system can automatically improve intelligibility by adjusting the speech speed to accommodate non-native speakers who are not proficient in the target language.
The working process of AIx Speed is illustrated in Fig.~\ref{fig:aix_speed}. 
The system first extracts the human voice from the target video. 
Next, it splits this voice into specified equal intervals. 
Next, using each segmented voice as input, the system calculates the optimal playback speed for each segment voice, taking into account the characteristics of the voice as a whole.
Finally, the system changes each voice to the specified playback speed and combines them into a single voice. 
At the same time, the combined voice is recognized by speech recognition to confirm that the resulting single voice is understandable. 
This system consists of two mechanisms, as shown in Fig.~\ref{fig:architecture}. 
One is a playback speed adjuster (left), and the other is a speech recognizer (right). 
The former is used to maximize the playback speed of the input speech, while the latter is used to evaluate how understandable the input speech is.
By training these two models simultaneously, it is possible to generate speech that plays back as fast as possible within the comprehension range.
The following subsections describe these two key features.

\begin{figure*}[t]
  \includegraphics[width=1.01\textwidth]{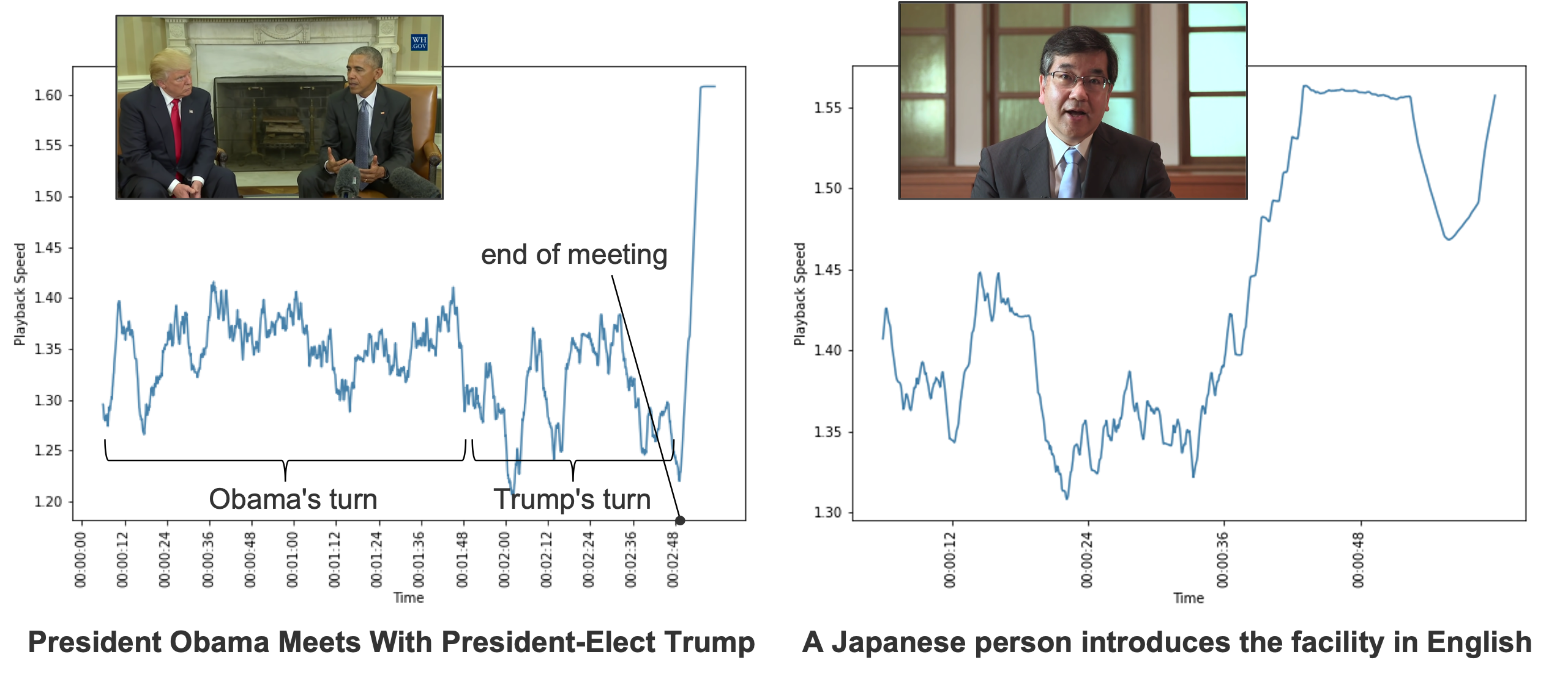}
  \caption{Application example. This graph shows how the playback speed changed when AIx Speed was applied to two videos on YouTube. As you can see, the playback speed changes flexibly according to the speaker's speech.}
  \Description{}
  \label{fig:application}
\end{figure*}

\subsection{Playback speed adjuster}
In this study, we use Wav2Vec2, a self-supervised neural network designed for speech signal processing systems, to optimize the playback speed. 
The playback speed controller is divided into a feature extractor layer and a linear layer, as shown in Fig.~\ref{fig:architecture}~(left). 
They are trained by pre-training through self-supervised representation learning on unlabeled speech data and regression learning, which outputs the playback speed based on the features after the representation learning. 
The pre-training method for the feature extractor layer is similar to masked language modeling, as exemplified by bidirectional encoder representations from transformers (BERT)~\cite{devlin-etal-2019-bert} in natural language processing, where a portion of the input is masked and the corresponding utterance features are estimated from the remaining input. 
In this way, the model can learn good-quality features of the target language to which the rate of utterance should be adapted. 
Typically, these self-supervised learners are used to tackle tasks such as speech recognition and speaker identification by pre-training and then fine-tuning with a small amount of label data. 
For example, in speech recognition, we have added a projection layer and a connectionist temporal classification (CTC) layer~\cite{Graves:06icml} to the output of self-supervised learners, such as Wav2Vec2 and HuBERT~\cite{Hsu2021}, to enable transcription from speech waveforms. 
Similar to these methods, we pre-train on unlabeled speech data and then connect and train a linear layer that outputs rates. 
In general speech processing tasks, such as speech recognition~\cite{Mishaim21} and speaker identification~\cite{Zx21, Kabir2021ASO}, the difference from the class label is minimized as an error function. 
On the other hand, the goal of the playback speed adjuster is to maximize the speed.
Therefore, we designed the following function whose value decreases as the speed increases.

\begin {align*}
    {\rm Loss_{speed}} = \left(\frac{1}{10}\right)^{\frac{1}{S}\sum_{s=1}^{S}r_s},
\end {align*}
where $r_t$ is the playback speed for each segment obtained using Wav2Vec2 and a linear layer with audio $X$ as input as follows:
\begin {align*}
  C = c_{\left[1:T\right]}= {\rm Wav2Vec2}\left(X\right), \\
  R = r_{\left[1:S\right]} = {\rm Linear}\left(C\right).
\end {align*}

\subsection{Speech recognizer}
The speech recognizer transcribes the speech converted to the playback speed obtained by the playback speed adjuster (Fig.~\ref{fig:architecture}~(right)). 
When speech parts with different playback speeds are combined, noise is generated in the speech data and the sound quality is degraded. 
We use voice separation technology~\cite{mcfee2015librosa} to extract only the speaker's voice and reduce the effect of noise. 
The resulting speech is then fed into a speech recognizer for speech recognition. The speech recognition process consists of the extraction of acoustic features from the speech waveform, estimation of the classes of acoustic features for each frame, and generation of hypotheses from the sequence of class probabilities. 
Since the speech recognizer can partially share the neural network with the playback speed adjuster, the overall network size can be reduced. 
As shown in Fig.~\ref{fig:architecture}, the dotted speech feature extractor is shared. 
The speech features obtained by this mechanism are used as input to generate text in the projection layer. 
In this process, the speech recognizer is trained to minimize CTC loss, as in normal speech recognition. 
In summary, the entire model is trained to minimize the following error function, which is a combination of this error function and the error function of the playback speed adjuster.
\begin {align*}
    {\rm Loss} = {\rm Loss_{speed}} + \lambda {\rm Loss_{ctc}}.
\end {align*}
Here, $\lambda$ is a hyperparameter that adjusts the importance of the playback speed calculation and speech recognition.

\section{Prototype}
As a prototype of AIx Speed, an application that optimizes audio playback speed using English as the target language has been implemented. 
This section describes the implementation and usage of the application.

\subsection{Implementation}
Wav2Vec2 was used for the prototype's shared utterance learning model (the utterance learning part shared by the speech recognizer and the playback speed adjuster).
For pre-training, LibriSpeech~\cite{Vassil2015} was used as the dataset, train-clean-360 as the training data, and dev-clean as the validation data.
The dataset consist of clean speech data in LibriSpeech and were partitioned using the data partitioning method proposed in LibriSpeech for training/validation.
The pre-training did not require the corresponding transcribed text, only the speech data.
We then trained a speech recognizer and a playback speed adjuster using two sets of speech data, including the transcribed text.
One was LibriSpeech's train-clean-100, and the other was the English speech database read by Japanese students (UME-ERJ)~\cite{Nobuaki2002}.
The latter is a dataset of English spoken by non-native Japanese speakers, with 202 speakers (100 males and 102 females) reading simple English sentences.
The playback speed adjuster and the speech recognizer were trained separately.
First, the speech recognizer was trained using LibriSpeech and UME-ERJ, and then the playback speed adjuster was trained using the same data with fixed weights for the speech recognizer.
All speech files used for training were normalized to a sampling frequency of 16 kHz.
The Wav2Vec2 used the initial parameters implemented in PyTorch~\cite{Adam2019}, and the final layers of both the playback rate adjuster and the speech recognizer were $768$-dimensional linear layers.
The hyperparameter $\lambda$ of the error function was set to $10^{-7}$, and training was performed for 5 epochs with a batch size of $32$ using the AdamW~\cite{loshchilov2018decoupled} optimization algorithm.

\begin{table*}[h]
  \caption{Performance comparison of models}
  \begin{minipage}{0.45\linewidth}
  \centering
  \subcaption{LibriSpeech}
  \begin{tabular}{cccc}
    Model & Avg. Speed & CER & WER \\ \hline
    Wav2Vec2 (1.00x) & 1.00 & \textbf{4.38} & \textbf{12.57} \\
    Wav2Vec2 (1.30x) & 1.30 & 5.61 & 14.58  \\
    Wav2Vec2 (1.50x) & 1.50 & 6.83 & 17.19  \\
    \textbf{AIx Speed} & 1.30 & \underline{5.21} & \underline{12.96} \\
  \end{tabular}
  \end{minipage}
  \begin{minipage}{0.45\linewidth}
  \centering
  \subcaption{UME-ERJ}
  \begin{tabular}{cccc}
    Model & Avg. Speed & CER & WER \\ \hline
    Wav2Vec2 (1.00x) & 1.00 & \underline{26.74} & \underline{55.02}  \\
    Wav2Vec2 (1.29x) & 1.29 & 33.71 & 63.90  \\
    Wav2Vec2 (1.50x) & 1.50 & 36.93 & 66.51 \\
    \textbf{AIx Speed} & 1.29 & \textbf{26.45} & \textbf{53.13} \\
  \end{tabular}
  \end{minipage}
  \label{table:technical_evaluation}
\end{table*}

\subsection{Usage of the application}
Examples of using AIx Speed are shown in Fig.~\ref{fig:application}. 
This is an example of the prototype applied to a video uploaded to YouTube. 
The horizontal axis is the playback time, and the vertical axis represents the playback speed output by the model for each playback time. The first is an example of speeding up a dialogue, movie, or lecture. 
Two speakers appear in the video, and the optimal playback speed can be set for each speaker. Of particular interest is that the two speakers in the video speak at different speeds, so the average playback speed for the two speakers is different. 
It can also be seen that the playback speed increases drastically from the moment when the dialog between the two speakers ends and there is no more speech from the person. 
Although we did not intend to design this feature, we can see that our system, like conventional playback speed controllers, can speed up the playback speed in the parts where there is no speech.
The second example is speeding up the speech of non-native speakers to make it easier to understand. 
Since the speech of non-native speakers is often slower than that of native speakers, moderately speeding up the speech makes it easier to understand. 
The change in playback speed shows that the overall playback speed is faster than the speech between native speakers in the first video. This indicates that the model can speed up the speech more because the non-native speakers' speech is slower than that of the native speakers.

\section{Evaluation}
\subsection{Technical evaluation}
To demonstrate that the proposed method can optimize playback speed while maintaining content understanding, we compared the CER and WER values at the AIx Speed--modified speech playback speed to those at a constant playback speed.
We compared the CER and WER with the speech playback speed modified by the AIx Speed to the CER and WER when the speech was simply played at a constant speed.
The speeds of the comparison targets were 1.0x, 1.5x and the average speed times the playback speed of AIx Speed.
A standard Wav2Vec2 based speech recognition model, which was the speech recognizer used in our method, was used to compute the CER and WER for comparison.
The performance of the models is shown in Table.~\ref{table:technical_evaluation}~(a) and \ref{table:technical_evaluation}~(b) for LibriSpeech and UME-ERJ, respectively.
AIx Speed produces speech 1.30 times faster on average for LibriSpeech and 1.29 times faster on average for UME-ERJ.
Both results show that the playback speed optimized by AIx Speed has lower values for both CER and WER than the average constant speech speed at that playback speed.
From these results, it can be said that the proposed model maximizes the playback speed while guaranteeing the recognition performance.
In addition, for UME-ERJ, the speech generated at AIx Speed shows better recognition performance in terms of WER than at 1.0x playback speed.
Therefore, it is also suggested that the proposed method can be used to convert the speech of non-native speakers into more understandable speech.

\subsection{User evaluation}
\begin{figure}[t]
  \includegraphics[width=0.48\textwidth]{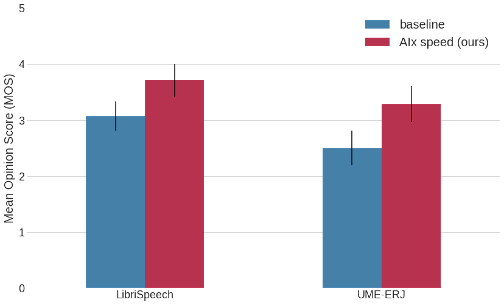}
  \caption{Voice quality at baseline (constant playback speed increase throughout) and AIx Speed (flexible playback speed increase)}
  \Description{}
  \label{fig:user_evaluation}
\end{figure}

User experiments were conducted to confirm that the generated speech was understandable.
The quality of the speech generated by the proposed method was compared with that of the speech played at a constant speed, at the average playback speed of the speech.
The quality was evaluated using the mean opinion score, which is commonly used in speech synthesis research~\cite{Aaron2016, Yuxuan2017}.
This measure rates speech quality on a five-point scale from 1 (poor) to 5 (excellent).
Participants were 50 US residents who used English on a daily basis.
20 sentences were extracted from each of the LibriSpeech and UME-ERJ datasets, and half were converted to speech with the proposed speed-up, while the other half were converted to speech with a constant speed-up.
Participants were given a total of 40 sentences of audio and assisted to rate the quality of the audio.
Figure~\ref{fig:user_evaluation} shows the quality of the generated speech at baseline and AIx Speed, LThe quality of LibreSpeech and UME-ERJ were 0.5 and 0.8 points higher at speeds generated by the proposed method, respectively.

\section{Discussion}
\subsection{Evaluation results}
The technical evaluation shows that the proposed method can produce speech that is easier to understand than that produced by simply increasing the playback speed in terms of speech recognition performance.
The user evaluation also shows that the proposed method can produce speech that is easier to understand for real users. These results show that the proposed method can produce speech with a high playback speed within a range that is easy for users to understand. 
This allows users to watch videos at a reasonable speed without having to adjust the playback speed for each video. 
However, the improvement in MOS values by using the proposed method is by no means sufficient.
In the current model, the average conversion to a faster playback speed is about 1.3 times, but it is a future task to investigate whether it is possible to make this even faster. 
In fact, many video playback services implement 1.5x and 2.0x playback speeds, and some people watch dramas and lectures at such speeds. Therefore, we expect that it will be possible to convert up to this speed and make the audio easy to understand. 
In addition, since each user has a different preferred playback speed, personalizing the model so that it plays at the optimal playback speed for users is also a future issue.

\subsection{Listening comprehension and speech recognition}
As discussed in the preliminary research chapter, it can be seen that there is a relationship between speech recognition performance and human transcription ability. 
Thus, we expect to build automated systems for various tasks and evaluations by replacing human speech comprehension ability with machine learning models, as in this system. 
On the other hand, speech recognition performance based on playback speed does not perfectly match human speech comprehension. 
In other words, as the playback speed increases, the dictation performance decreases in both cases, but the performance values are not exactly the same.
Thus, we anticipate that by training speech recognition models to match these relationships as closely as possible, it will be possible to use them more generally as alternatives to humans. Distillation, a technique that learns to approximate an output that matches existing results, will be the technical key.

\subsection{Adjustment of non-native speakers' speech}
Several suggestions can be made as to how increasing the playback speed by the proposed system improves the intelligibility of speech for non-native speakers. One of them is that when non-native speakers read English manuscripts, they may find it easier to understand if they speak naturally (slower from a native speaker's point of view) and then artificially speed up their speech, rather than forcing them to speak quickly like a native speaker. In fact, this study also began with the realization that it is easier to listen to a video of a non-native speaker speaking his or her native language when it is played at a high speed.

\section{Conclusion}
This paper presents a system that applies a speech recognition model to automatically and flexibly adjust the playback speed of video and audio within the range of human comprehension.
By using this system, users can consume audiovisual content at optimal speeds without having to manually adjust the playback speed.
Experiments have also confirmed that the system makes it easier for users to understand the speech of non-native speakers.
In the future, we expect the system to be used in a variety of applications, such as video distribution services and language learning tools.

\begin{acks}
  This work was supported by JST Moonshot R\&D Grant Number JPMJMS2012, JST CREST Grant Number JPMJCR17A3, and The Univesity of Tokyo Human Augmentation Research Initiative.
\end{acks}

\bibliographystyle{ACM-Reference-Format}
\bibliography{reference}

\end{document}